\documentclass[10pt, a4paper]{article}

\usepackage[final]{lrec2026} 

\usepackage{float}
\usepackage{amsmath}
\usepackage{amssymb}
\usepackage{booktabs}
\usepackage{subcaption}
\usepackage{diagbox}
\usepackage{xcolor}
\usepackage{pifont} 
\usepackage[most]{tcolorbox} 
\tcbuselibrary{breakable}    

\usepackage{soul}
\sethlcolor{yellow}

\newcommand{\cmark}{\textcolor{green!60!black}{\ding{51}}} 
\newcommand{\xmark}{\textcolor{red}{\ding{55}}}        

\usepackage{xcolor}  

\usepackage{pbox}
\usepackage{makecell}
\usepackage{multirow}
\usepackage{graphicx}

\newcommand*\colourcheck[1]{%
  \expandafter\newcommand\csname #1check\endcsname{\textcolor{#1}{\ding{51}}}%
}
\newcommand*\colouruncheck[1]{%
  \expandafter\newcommand\csname #1uncheck\endcsname{\textcolor{#1}{\ding{53}}}%
}
\colourcheck{green}
\colouruncheck{red}
\colouruncheck{blue}

\newcommand{\update}[1]{\textcolor{black}{#1}}
\newcommand{\numedit}[1]{\textcolor{black}{#1}}

\title{SciClaimEval: Cross-modal Claim Verification in Scientific Papers}

\name{Xanh Ho,$^1$ 
Yun-Ang Wu,$^{2, 3}$ 
Sunisth Kumar,$^4$ 
Tian Cheng Xia,$^{5*}$  \and 
Florian Boudin,$^6$ 
Andr\'{e} Greiner-Petter,$^{1,7}$ 
and Akiko Aizawa,$^{1, 3, 4}$ 
}

\address{
$^1$National Institute of Informatics, Japan \hspace{1cm}
$^2$National Taiwan University \hspace{1cm} $^3$NII LLMC, Japan \addrbreak
$^4$The University of Tokyo, Japan \hspace{1cm}
$^5$University of Bologna, Italy \addrbreak
$^6$Inria, LS2N, Nantes Université, France \hspace{1cm}
$^7$University of Göttingen, Germany \addrbreak
\{xanh, yunangwu, aizawa\}@nii.ac.jp, sunisth@g.ecc.u-tokyo.ac.jp \addrbreak
tiancheng.xia@studio.unibo.it, florian.boudin@univ-nantes.fr, greinerpetter@gipplab.org 
}

\abstract{
We present SciClaimEval, a new scientific dataset for the claim verification task. Unlike existing resources, SciClaimEval features authentic claims, including refuted ones, directly extracted from published papers.
To create refuted claims, we introduce a novel approach that modifies the supporting evidence (figures and tables), rather than altering the claims or relying on large language models (LLMs) to fabricate contradictions. The dataset provides cross-modal evidence with diverse representations: figures are available as images, while tables are provided in multiple formats, including images, LaTeX source, HTML, and JSON. SciClaimEval contains 1,664 annotated samples from 180 papers across three domains, machine learning, natural language processing, and medicine, validated through expert annotation. We benchmark 11 multimodal foundation models, both open-source and proprietary, across the dataset. Results show that figure-based verification remains particularly challenging for all models, as a substantial performance gap remains between the best system and human baseline.
 \\ \newline  }

\begin{document}

\maketitleabstract

\section{Introduction}

\begingroup\def\thefootnote{*}\footnotetext{Research conducted during internship at NII, Japan.}\endgroup

Scientific claim verification involves determining whether claims made in research papers are supported or refuted by the accompanying evidence~\cite{wadden-etal-2020-fact,guo-etal-2022-survey}.
With the rapid rise of generative AI and large language models (LLMs), the number of submissions to scientific conferences and journals has grown substantially, creating a greater demand for tools that help reviewers assess the validity of authors’ claims. 
%
Accurate and reliable claim verification systems could significantly strengthen the peer-review process by automatically identifying unsupported or inconsistent claims.
However, the ability of existing models, including recent multimodal LLMs (MLLMs), to perform this task remains largely unexplored due to the lack of a comprehensive dataset.

Over the past few years, several datasets have been introduced for claim verification in scientific papers, including SciFact~\cite{wadden-etal-2020-fact}, PubHealth~\cite{kotonya-toni-2020-explainable-automated}, SciTab~\cite{lu-etal-2023-scitab}, and more recently, SciVer~\cite{wang-etal-2025-sciver}. 
SciFact and PubHealth are text-only datasets in the medical domain, while SciTab focuses on computer science and provides tables as evidence.
SciVer extends this line of work by incorporating multimodal evidence, while remaining focused on computer science. 
However, SciVer has two key limitations: first, its claims are synthetic, written by experts rather than drawn from actual papers, and second, its tables are provided solely as images, without corresponding structured formats.
More broadly, a persistent limitation across most existing datasets lies in how unsupported claims are constructed.
These are often created by editing supported claims to introduce contradictions with the evidence, which can lead to artifacts and heuristic cues.
For example, many refuted claims simply rely on inserting negation words such as not, making the task less realistic.

In this paper, we present SciClaimEval, a new dataset for cross-modal scientific claim verification. 
Our work differs from existing resources in three important ways.
First, all claims, both supported and refuted, are authentic, sourced directly from published papers across three domains: machine learning (ML), natural language processing (NLP), and medicine. 
Second, unlike prior work that generate refuted claims by altering original statements or relying on LLMs, we introduce a novel strategy that creates negative examples by disturbing the supporting evidence itself, specifically by modifying figures and tables. 
%
Third, SciClaimEval provides cross-modal evidence with rich and diverse representations: both tables and figures are included, and tables are available in multiple formats, including images, LaTeX source, HTML, and JSON.

\begin{table*}[ht]
  \begin{center}
  \resizebox{\textwidth}{!}{%

\begin{tabular}{c l c l l l c l l c}
\toprule
\textbf{Year} & \textbf{Dataset} & \textbf{Size} & \textbf{Domain} &
\textbf{\makecell[l]{Fact Source \\ (Positive)}} & \textbf{\makecell[l]{Fact Source \\ (Negative)}} &
\textbf{\makecell[l]{Authentic \\ Claims}} & \textbf{\makecell[l]{Input \\ Context}} & \textbf{Sources} & \textbf{Verdict} \\

\midrule

2021 & SEM-TAB-FACTs & 5,715 & Multi & Crowd & Crowd & \xmark \xmark & A table (.xml) & ScienceDirect & 3 labels \\

2023 & SciTab & 1,224 &
CS &
\makecell[l]{Original \\ sentences} &
Claim modification &
\cmark \xmark & A table (.json) & SciGen & 3 labels \\

2025
& MMSci-Eval & 3,114 &
CS &
\makecell[l]{GPT followed \\ by verification} &
Claim modification &
\xmark \xmark & A table image & SciGen & 3 labels \\

2025 & SciAtomicBench & 2,568 &
Multi &
\makecell[l]{GPT followed \\ by verification} &
Claim modification &
\xmark \xmark & A table (.json) &
\makecell[l]{SciGen, \\ PubTables-1M, \\ Financials, \\ MatSciTable} &
2 labels \\

2025 & SciVer & 3,000 &
CS &
Experts &
Claim modification &
\xmark \xmark &
\makecell[l]{Textual paras, \\ multiple charts, \\multiple tables \\ (table image)} &
arXiv papers & 2 labels \\

2025 & MuSciClaims & 1,515 &
Multi & \makecell[l]{Original \\ sentences} & 
Claim modification &
\cmark \xmark & A figure &
\makecell[l]{Nature Physics, \\ Journal of \\ the American \\Chemical Society, \\ and Cell} &
3 labels \\

\midrule 
2025 & \makecell[l]{SciClaimEval \\ (ours)} & 1,664 & Multi & \makecell[l]{Original \\ sentences} & Evidence modification & \cmark \cmark & \makecell[l]{A table* or \\ a figure} & arXiv, PeerJ & 2 labels \\

\bottomrule
\end{tabular}

}
\caption{Comparison of SciClaimEval with existing multimodal scientific claim verification datasets.
For the Authentic Claims column, the two bars from left to right represent a supported claim and a refuted claim, respectively. The asterisk (*) next to Table indicates that our table is available in multiple formats, including table image, JSON, and HTML. 
}
\label{sec_related_dataset}
  \end{center}
\end{table*}

Specifically, SciClaimEval is constructed in three main steps. 
First, we collect a pool of papers spanning three domains: NLP, ML, and medicine.
For the medical domain, we use papers from PeerJ, while for NLP and ML, we gather papers from arXiv, referencing accepted papers from the ACL Anthology and NeurIPS as guidance.
For each paper, we extract the main text, figures, and tables. 
Second, we perform claim–evidence pair extraction. 
%
Using keywords such as ``Table 1'' or ``Tab. 1'', we identify sentences that explicitly reference a table or figure and pair each of them with the corresponding evidence.
Third, we conduct expert annotation. 
All claim-evidence pairs are reviewed by experts, who carry out two tasks: 1) claim-evidence verification and 2) evidence modification. 
When a claim is annotated as supported, the annotators modify the corresponding evidence to create an unsupported version.
In total, SciClaimEval contains \numedit{1,664} claims from \numedit{180} papers, covering both figures and tables across three domains.


We then evaluate 11 multimodal foundation models on SciClaimEval, covering both open-source and proprietary models of varying sizes. Our results show that the figure-based subset is challenging for all models, including o4-mini, with a substantial gap remaining between the best model performance and the human baseline. In contrast, the table-based subset is primarily useful for assessing open-source MLLMs, as o4-mini performs close to the human baseline. Additionally, our table data includes diverse formats, providing a valuable resource for future research on processing scientific papers.\footnote{Our dataset is available at \url{https://sciclaimeval.github.io/}}

\section{Related Work}
In this section, we first review studies on general-domain claim verification, focusing on text-based approaches, and then discuss research related to multimodal scientific claim verification.

\paragraph{Claim Verification.}
Claim verification (fact-checking) has long been a central problem in NLP and AI, with extensive progress surveyed by~\citet{guo-etal-2022-survey}. 
The task aims to determine whether a claim is supported, refuted, or unverifiable given available evidence.
Research on general-purpose, text-based verification has primarily focused on two sources: news articles and Wikipedia. In the news domain, several benchmark datasets have been developed, including LIAR~\cite{wang-2017-liar} and MultiFC~\cite{augenstein-etal-2019-multifc}. 
In the Wikipedia domain, widely used datasets include FEVER~\cite{thorne-etal-2018-fever} and HoVer~\cite{jiang-etal-2020-hover}. Both datasets contain claims that often require reasoning over multiple documents for verification.

Beyond purely textual settings, the community has also introduced resources for verification over structured and multimodal evidence. These include tables (TabFact~\cite{Chen2020TabFact} and  FEVEROUS~\cite{aly2021feverous}), figures and charts (ChartCheck~\cite{akhtar-etal-2024-chartcheck}), and knowledge graphs (FactKG~\cite{kim-etal-2023-factkg}), broadening the scope of reasoning beyond unstructured text.

In parallel, domain-specific efforts have targeted scientific claims, where veracity judgments depend on rigorously sourced scholarly evidence. Notable datasets include SciFact~\cite{wadden-etal-2020-fact}, PubHealth~\cite{kotonya-toni-2020-explainable-automated}, and SciFact-Open~\cite{wadden-etal-2022-scifact}.

\begin{figure*}[h]
\centering
    \includegraphics[width=0.96\textwidth,trim={6.7cm 6.8cm 8.3cm 4.8cm},clip]{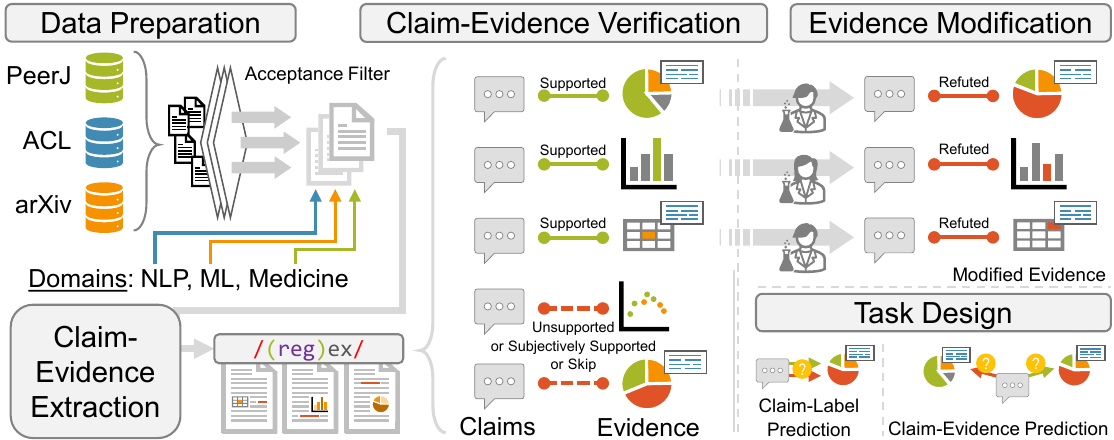}
    \caption{Our dataset construction pipeline consists of three main steps: data preparation, automatic claim-evidence extraction, and human annotation (Subsections~\ref{subsec_data_prepare},~\ref{subsec_claim_extract}, and~\ref{subsec_human_annotate}). The human annotation process involves two tasks: claim-evidence verification and evidence modification.
    \update{After collecting all samples, we design two subtasks in our dataset: claim-label prediction and claim-evidence prediction. Details are in \ref{subsec_task_design}.}
    }
    \label{fig_overall}
\end{figure*}

\paragraph{Multimodal Scientific Claim Verification.}
Unlike the datasets presented in the previous section, this section discusses datasets that involve multimodal scientific content.
Table~\ref{sec_related_dataset} provides a comparison of existing datasets and our dataset.
SciTab~\cite{lu-etal-2023-scitab}, SEM-TAB-FACTs~\cite{wang-etal-2021-semeval}, and SciAtomicBench~\cite{zhang2025atomicreasoningscientifictable} are datasets that focus on using tables from scientific papers. The tables in SciTab and SciAtomicBench are represented in JSON format, whereas the tables in SEM-TAB-FACTs are represented in XML format.
MMSci-Eval~\cite{yang2025doestablesourcematter} also uses tables as evidence, but the tables are represented as images.
MuSciClaims~\cite{lal-etal-2025-musciclaims} focuses on claim verification using figures; their figures are often complex and contain multiple subcharts and tables.
SciVer~\cite{wang-etal-2025-sciver} emphasizes the use of multiple charts, tables, or textual paragraphs as evidence, highlighting reasoning over multiple pieces of information.
A limitation of all existing datasets is that they only modify claims to create refuted examples. These modifications typically involve generating the opposite meaning or performing semantic flips of existing claims, which can lead to spurious patterns and shortcut reasoning.

Another key aspect of our dataset is that both supported and refuted claims are authentic claims extracted directly from scientific papers.
While datasets such as SciTab and MuSciClaims also reuse authentic claims from existing papers, they do so only for supported claims, limiting the diversity of refuted examples.
With AI increasingly assisting in scientific writing and reviewing, using authentic claims is crucial to reflect real-world challenges in scientific reasoning.

\section{Dataset Construction and Analysis}
\label{sec_dataset_construction}

In this section, we first describe our dataset construction process, which consists of three main steps: data preparation, automatic claim and evidence extraction, and human annotation, corresponding to Subsections~\ref{subsec_data_prepare},~\ref{subsec_claim_extract}, and~\ref{subsec_human_annotate}.
\update{
We then provide detailed information about the resulting dataset, including dataset statistics, dataset analysis, dataset validation, and task design, presented in Subsections~\ref{subsec_data_statistics},~\ref{subsec_data_analysis},~\ref{subsec_data_validate}, and~\ref{subsec_task_design}.
}
Figure~\ref{fig_overall} illustrates the overall process of our dataset construction.

\subsection{Data Preparation}
\label{subsec_data_prepare}

In our dataset creation process, we collect papers from three sources: PeerJ for the medical domain, ACL for NLP, and AI/ML conferences for machine learning. 
For NLP domain, papers are collected from the ACL Anthology and manually mapped to their arXiv versions when available. We select papers from a variety of conferences, including but not limited to EMNLP, ACL, EACL, NAACL, and IJCNLP.
For AI/ML domain, we select arXiv papers that include comments indicating acceptance at AI/ML conferences, such as NeurIPS.
The arXiv IDs from both the NLP and AI/ML collections are used to retrieve data from ar5iv~\cite{SML:ar5iv:04:2024}, which provides HTML-rendered versions of arXiv papers.
The main text of each paper is cleaned and parsed into JSON format, storing the title, abstract, and a list of paragraphs for each section. Figures are extracted from ar5iv and saved in PNG format. To obtain high-quality table data, we additionally download the LaTeX source of each arXiv paper and extract tables directly from the source files, preserving them in LaTeX format.
For medical domain, we select papers published in PeerJ between 2024 and 2025 in the Medicine Articles category. Since PeerJ provides HTML versions of papers, we crawl the raw HTML and parse it into JSON. Figures and tables are extracted separately, with tables stored in HTML format and figures in PNG.

\subsection{Claim-Evidence Extraction}
\label{subsec_claim_extract}

We use regular expressions to identify paragraphs in each paper that mention either a table or a figure. When such a reference is found, we extract the corresponding paragraph, split it into individual sentences, and pair each sentence with the mentioned table or figure, treating it as supporting evidence.
By applying this process to all papers, we construct a set of sentence–evidence pairs, where the evidence corresponds to either a table or a figure.
Our dataset focuses solely on the main text of the papers, excluding any content from appendices, including their tables and figures.
Following the findings of~\citet{ho-etal-2025-table}, which suggest that a lack of context can lead to task ambiguity, we also include the preceding sentences from the same paragraph as a short contextual field for each claim sentence.

\subsection{Human Annotation}
\label{subsec_human_annotate}

Our human annotation process includes two tasks: claim-evidence verification and evidence modification. We first describe the annotators, then detail the annotation process.

\paragraph{Annotators Information.}
We recruited annotators from the students of our laboratory. Including the authors of this paper, we have a total of 11 annotators, all of whom are either graduate students or expert researchers in the NLP and ML domains.
We did not provide direct monetary compensation; however, the annotators were rewarded with a fully funded travel trip covering all associated costs.

\paragraph{Claim-Evidence Verification.}
From the list of extracted claim–evidence pairs for each paper, annotators are presented with a claim, its corresponding evidence (either a figure or a table), and the preceding sentences from the same paragraph as short context. If the claim is the first sentence, the context is empty.
Annotators are asked to:
\textit{\textbf{1. Assign a label}}: Choose one of the following: \textbf{Supported} (evidence clearly supports the claim with no ambiguity), \textbf{Subjectively Supported} (the claim contains subjective terms such as ``large margin'' or ``competitive,'' making it difficult for annotators to determine whether these adjectives are accurate), \textbf{Unsupported} (evidence does not support the claim), or \textbf{Skip} (insufficient knowledge to judge, or the claim is problematic or purely descriptive).
\textit{\textbf{2. Indicate context use}}: Choose one of the following: No (the claim is understandable without context), Yes (the short context is needed), or Other sources (the full paper is needed to understand the claim).

\paragraph{Evidence Modification.}
If a claim–evidence pair is labeled as \textit{Supported}, annotators proceed to the evidence modification task. 
We use the same input as in the previous task and provide annotators with a list of modification operations, including explanations and examples. 
The goal is to modify the table or figure so that the claim becomes \textit{Unsupported} when paired with the altered evidence. 
%
%
For tables, annotators can perform the following operations:
\begin{itemize}
    \item \textbf{Change the cell values:} Modify the content of one or more cells.
    \item \textbf{Swap rows or columns:} Move the name of a row or column to another one.
    \item \textbf{Alter the table:} Add or remove rows or columns.
    \item \textbf{Others:} Annotators may introduce additional changes.
\end{itemize}

For figures, annotators can perform the following operations:
\begin{itemize}
    \item \textbf{Graph Flip:} Flip a graph or part of it.
    \item \textbf{Legend Swap:} Swap text in the legend.
    \item \textbf{Graph Swap:} Exchange graphs or subgraphs from the same paper.
    \item \textbf{Category Swap:} Swap category labels to contradict the claim.
    \item \textbf{Others:} Propose other types of modifications as appropriate for the sample.
\end{itemize}

In both cases, when annotators select \textbf{Others}, they are required to record the details of the changes.

\begin{table}[h!]
    \centering
    \begin{tabular}{l r r r}
\toprule
\textbf{Property} & \textbf{Val} & \textbf{Test} & \textbf{All Data} \\
\midrule
\multicolumn{4}{l}{\textit{Labels}} \\ 
\#Supported & 395 & 481 & 876 \\
\#Refuted & 352 & 436 & 788 \\

\midrule
\multicolumn{4}{l}{\textit{Modality Type}} \\ 
Table & 482 & 523 & 1,005\\
Figure & 265 & 394 & 659 \\

\midrule
\multicolumn{4}{l}{\textit{Context Use}} \\ 
No & 494 & 619 & 1,113\\
Short Context & 148 & 190 & 338\\
Full Paper & 105 & 108 & 213 \\

\midrule
\multicolumn{4}{l}{\textit{Domain}} \\ 
NLP & 388 & 389 & 777 \\
ML & 162 & 240 & 402 \\
Medicine & 197 & 288 & 485\\

\midrule
\#Papers & 139	&164	& 180 \\
Total examples & 747 & 917 & 1,664 \\
\bottomrule
\end{tabular}






    \caption{Dataset statistics of SciClaimEval.}
    \label{tab_dataset_stats}
\end{table}

\subsection{Dataset Statistics}
\label{subsec_data_statistics}

The statistics of our dataset are presented in Table~Table~\ref{tab_dataset_stats}. We report the number of claims in the validation and test sets, as well as for the entire dataset. Information is provided across different properties, including modality type (table or figure), context usage, and domain. We split the data into validation and test sets based on context use and the type of operation from the evidence modification task. All \textit{Others} operations are included in the test set. Additionally, for tables, we retain the \textit{Alter the table} operations and some \textit{Swap rows or columns} operations in the test set.

\paragraph{Supported Claim Only.}
As shown in Table~\ref{tab_dataset_stats}, we have \numedit{876} supported claims but only \numedit{788} refuted claims.
In theory, the numbers of supported and refuted claims should be the same, since each evidence table or figure is modified to create an unsupported claim using the altered evidence.
However, during the dataset annotation process, we found that in some cases it was very difficult to modify the evidence in a way that remained logical and still posed a meaningful challenge to the models, rather than simply making superficial changes.
As a result, we have \numedit{88} ``supported claim only'' samples, \numedit{67} of which come from figure-based evidence and \numedit{21} from table-based evidence.

\paragraph{Different Formats of Table Evidence.}
For figure evidence, the figures are provided in .png format.
For table evidence, the original tables in the NLP and ML domains are in LaTeX format, while the tables in the Medicine domain are in .html format.
We also obtain both .png and .json versions of the table data.
To generate the .png format, the LaTeX files are first compiled into PDFs, which are then converted to PNG images using the Python pdf2image library~\cite{belval_pdf2image}. This library relies on built-in system tools for PDF manipulation.
The HTML files are converted directly into images using wkhtmltoimage~\cite{wkhtmltopdf}, an open-source tool that renders HTML files into various formats.
To estimate the accuracy of our table evidence in .png format, we randomly select 100 samples and manually evaluate them. 
We find that only one sample is rendered incorrectly because of overlapping columns.
Additionally, three samples have issues with the caption, as two have incomplete captions and one has no caption.

To generate JSON from LaTeX files, we first tried a rule-based approach but found it inadequate due to numerous edge cases. We then adopted GPT-5-nano to convert LaTeX and HTML tables into a predefined JSON format following the schema of~\citet{lu-etal-2023-scitab}, which includes the table ID, caption, column names, and cell values.
To estimate the accuracy of our table evidence in .json format, we randomly select 100 samples and manually evaluate them. 
We found 22 cases with minor issues (e.g., missing the top row due to multicolumns) and 19 cases with major issues where the JSON table content did not match the original table.

\begin{table}[h!]
    \centering
     \resizebox{0.85\columnwidth}{!}{%
    \begin{tabular}{l r r r}
\toprule
\textbf{Information} & \textbf{Max} & \textbf{Min} & \textbf{Avg.} \\

\midrule

Claim length & 91 & 7 & 25.6 \\

Table caption length & 134 & 3 & 29.9 \\

Figure caption length & 206 & 7 & 53.4  \\

Context length & 296 & 8 & 67.7  \\

\bottomrule
\end{tabular}
    }
    \caption{Detailed analyses of text lengths (based on word count) in SciClaimEval.}
    \label{tab_dataset_analysis}
\end{table}

\begin{figure}[h]
    \centering
    \includegraphics[width=1.02\columnwidth]{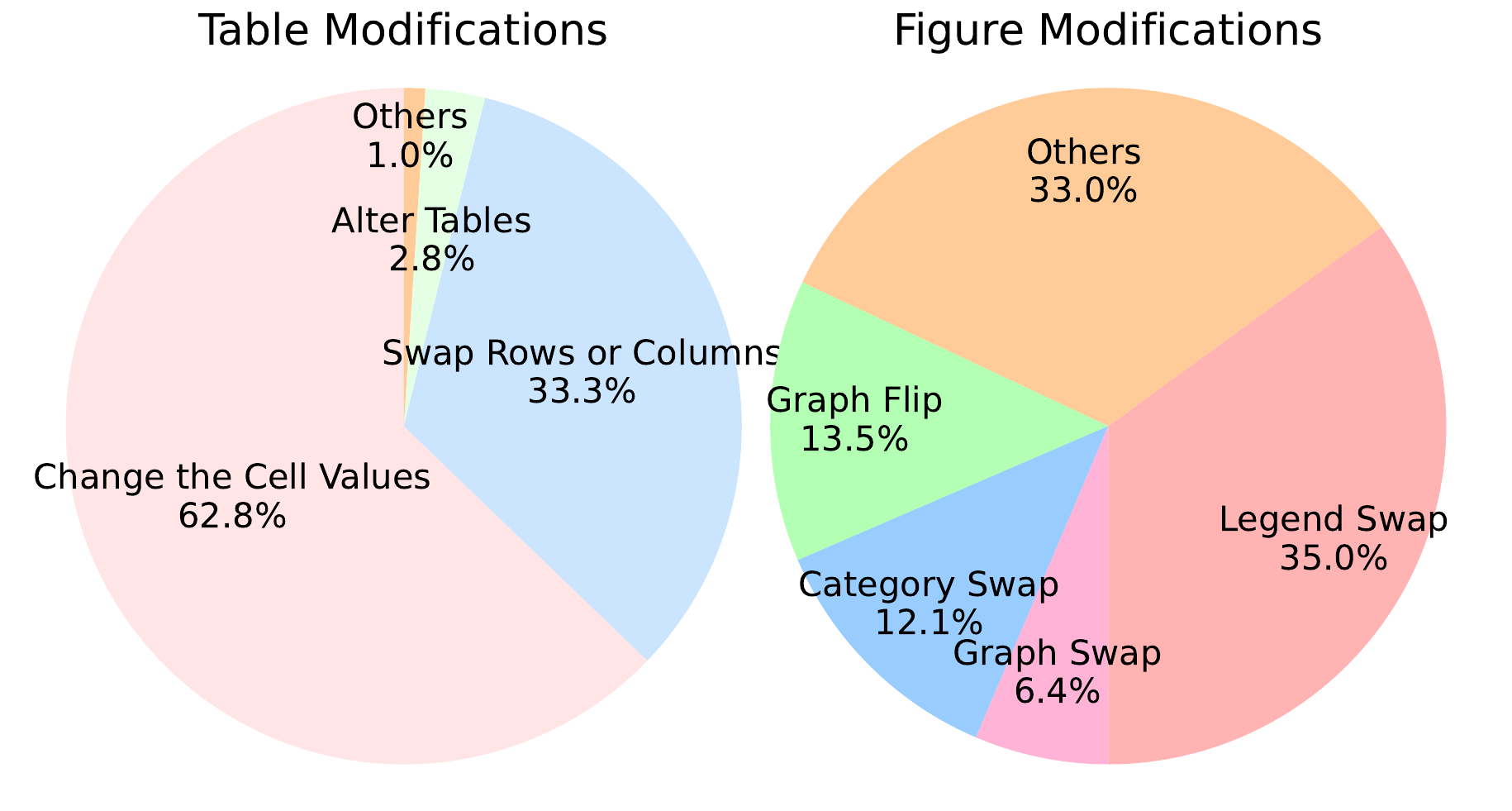}
    \caption{Analyses of evidence-modifying operations in SciClaimEval.}
    \label{fig_operation}
\end{figure}

\begin{figure*}[h]
\centering
    \includegraphics[width=0.95\textwidth,trim={0px 2px 4px 0px},clip]{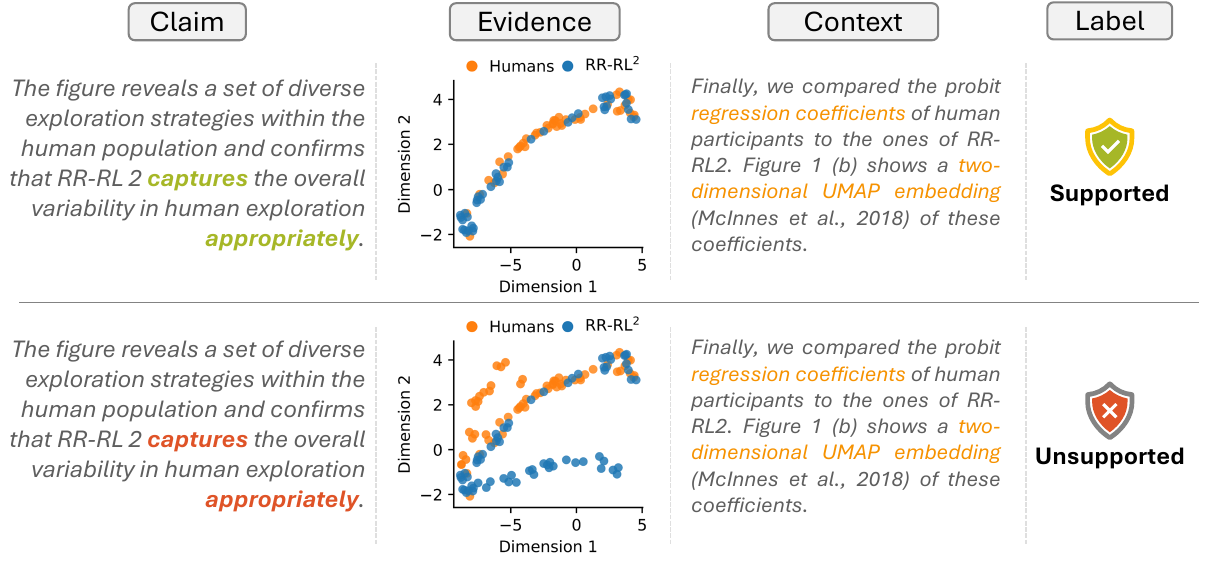}
    \caption{
    An example of an \emph{others} modification in the figure evidence from our dataset involves creating an unsupported claim by adding spurious data points. The annotator labels this operation as \emph{others}, with the specific detail noted as ``adding fake data points.'' The context provides the necessary information to understand the plot in the evidence.
    }
    \label{fig_example}
\end{figure*}

\subsection{Dataset Analysis}
\label{subsec_data_analysis}

\paragraph{Text Lengths.}
Table~\ref{tab_dataset_analysis} presents detailed analyses of the word counts for claims, table captions, figure captions, and short contexts.
As shown in the table, figure captions are generally longer than table captions.
On average, claims contain 25.6 words, ranging from 7 to 91 words.

\paragraph{Evidence-Modifying Operations.}
We present the operation analyses for table evidence and figure evidence modifications in Figure~\ref{fig_operation}.
As illustrated in the figure, the most common modification for table evidence is \emph{changing cell values}, while for figure evidence, \emph{legend swapping} is the most frequent. Annotators appear to be more creative when working with figure evidence, as they often select \emph{others} as the operation type. In contrast, \emph{others} is rarely chosen for table evidence.
Upon examining the details categorized as \emph{others} in the figure evidence, we find that annotators frequently perform operations such as \emph{changing bar heights}, \emph{manipulating data points} (e.g., moving or adding fake data points), \emph{adjusting axes}, or \emph{rearranging graphs}.
Figure~\ref{fig_example} shows an example in which the annotator labels the operation as \emph{others}, with the specific detail noted as ``adding fake data points.''

\begin{figure}[h]
    \centering
    \includegraphics[width=1\columnwidth]{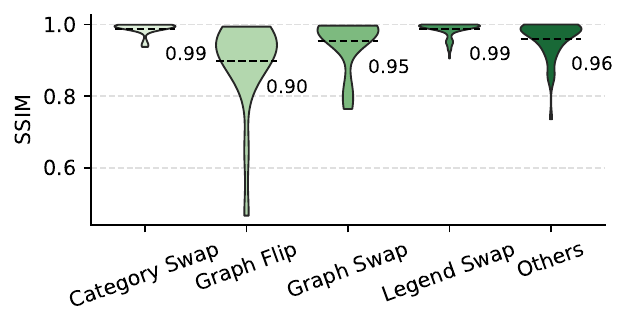}
    \caption{
        Violin plots showing the distribution of the Structural Similarity Index~\cite[SSIM;][]{wang2004ssim} across five operation types. 
        In each plot, the black dashed line indicates the group mean, and the numeric label to the right of the line denotes the corresponding average value. 
    }
    \label{pixel-changes}
\end{figure}

\paragraph{Pixel Changes in Figure Evidence Modification.}

Figure~\ref{pixel-changes} illustrates the image similarity between the original and modified images for each operation type. The Structural Similarity Index~\cite[SSIM;][]{wang2004ssim} was used as the metric for image similarity. 
We exclude 11 pairs with scaled or misaligned edits, as such cases would heavily skew the SSIM score.
We observe that \emph{graph flip} and \emph{graph swap} result in the most significant changes, which is reasonable since these two operations modify large areas of the chart. In contrast, \emph{category swap} and \emph{legend swap} produce minimal changes, as altering categories or legends is typically a localized operation. The \emph{others} category lies between these two groups, as it contains a variety of mixed operations.


\subsection{Dataset Verification}
\label{subsec_data_validate}

To establish a baseline for human performance on the dataset, we evaluated 80 samples, including 45 with table evidence and 35 with figure evidence.
Each subset was independently annotated by two annotators, resulting in a total of four annotators.
The annotators were graduate students and AI researchers.
On the table subset, the average macro-F1 score was 87.9, with an inter-annotator agreement of 86.7\%.
For the figure subset, the average macro-F1 score was 89.6, with an agreement rate of 91.4\%.

\subsection{Task Design}
\label{subsec_task_design}
\update{
After collecting all samples, we design two subtasks in our dataset.
The first subtask is claim-label prediction, which is the main task and follows prior work on claim verification datasets. 
The input consists of a claim, an evidence file (which can be a figure or a table) along with its caption, and additional contextual information intended to reduce ambiguity in the sample. The output is a label indicating whether the claim is supported or refuted.
The second subtask is claim-evidence prediction, whose goal is to identify which piece of evidence supports a given claim. This task is particularly challenging because the two evidence files are highly similar, making it difficult to distinguish the correct supporting evidence.
}

\begin{table*}[htb!]
    \centering
    \resizebox{\textwidth}{!}{%

\begin{tabular}{l | rrrrrr | rrrrrr}
    \toprule
    \multirow{3}{*}{\textbf{Model}} &

    \multicolumn{6}{c}{\textbf{Validation}} &
    \multicolumn{6}{c}{\textbf{Test}}  \\ 

    \cmidrule(rl){2-7}
    \cmidrule(rl){8-13}

   & \multicolumn{2}{c}{\textbf{No Context}} &
    \multicolumn{2}{c}{\textbf{Use Context}} &  
    \multicolumn{2}{c}{\textbf{Average}} &  
    \multicolumn{2}{c}{\textbf{No Context}} &
    \multicolumn{2}{c}{\textbf{Use Context}} &  
    \multicolumn{2}{c}{\textbf{Average}}  \\

    \cmidrule(rl){2-3}
    \cmidrule(rl){4-5}
    \cmidrule(rl){6-7}
    \cmidrule(rl){8-9}
    \cmidrule(rl){10-11}
     \cmidrule(rl){12-13}

    ~ & F1 & P-Acc & F1 & P-Acc & F1 & P-Acc & F1 & P-Acc & F1 & P-Acc & F1 & P-Acc \\\midrule

    \#Samples & 494 & 231 & 253 & 121 & 747 &	352 & 619 & 294 & 298 & 142 & 917	& 436\\

    \midrule
    llava-mistral-7b & 45.9 & 1.3 & 49.8 & 2.5 & 47.9 & 1.7 & 47.4 & 2.4 & 49.7 & 0.7 & 48.4 & 1.8 \\ 
    
    llava-vicuna-13b & 25.8 & 0.0 & 27.2 & 0.8 & 26.3 & 0.3 & 25.9 & 0.3 & 29.5 & 0.0 & 27.1 & 0.2 \\ 

    \midrule
    Llama-3.2-11B-Vision & 49.0 & 11.7 & 47.8 & 9.1 & 48.6 & 10.8 & 50.1 & 17.3 & 44.8 & 8.5 & 48.5 & 14.4 \\ 

    \midrule
    InternVL3\_5-1B & 52.3 & 21.6 & 52.3 & 15.7 & 52.3 & 19.6 & 51.6 & 22.4 & 52.1 & 16.9 & 51.8 & 20.6 \\
    
    InternVL3\_5-8B & 68.5 & 40.7 & 59.0 & 24.8 & 65.4 & 35.2 & 66.4 & 38.8 & 55.2 & 26.1 & 63.0 & 34.6 \\ 
    
    InternVL3\_5-14B & 70.8 & 46.3 & 63.7 & 32.2 & 68.5 & 41.5 & 68.0 & 40.5 & 59.3 & 27.5 & 65.3 & 36.2 \\ 
    
    InternVL3\_5-38B & 70.8 & 45.0 & 61.8 & 30.6 & 67.8 & 40.1 & 70.1 & 45.6 & 64.0 & 34.5 & 68.2 & 42.0 \\ 

    \midrule
    Qwen3-VL-4B & 71.5 & 46.3 & 68.7 & 38.8 & 70.6 & 43.8 & 70.4 & 45.6 & 67.5 & 36.6 & 69.6 & 42.7 \\ 
    
    Qwen3-VL-8B & 72.2 & 47.6 & 71.3 & 45.5 & 72.1 & 46.9 & 70.5 & 46.6 & \underline{69.1} & \underline{39.4} & 70.1 & 44.3 \\ 
    
    Qwen3-VL-30B-A3B & \underline{76.2} & \underline{55.0} & \underline{75.5} & \underline{54.5} & \underline{76.0} & \underline{54.8} & \underline{73.5} & \underline{49.7} & 67.0 & \underline{39.4} & \underline{71.4} & \underline{46.3} \\ 

    \midrule 
    o4-mini & \textbf{82.8} & \textbf{68.0} & \textbf{83.1} & \textbf{68.6} & \textbf{82.9} & \textbf{68.2} & \textbf{80.3} & \textbf{63.3} & \textbf{76.5} & \textbf{54.9} & \textbf{79.1} & \textbf{60.6} \\

    


    
    
    

    
    


\bottomrule

\end{tabular}

    }
    \caption{
    Macro-F1 (denoted as \textit{F1} in the table) and pair accuracy (denoted as \textit{P-Acc} in the table) of the models on our dataset.
    For the \textit{F1} columns, the number in the \textit{Samples} row represents the number of individual samples, whereas for the \textit{P-Acc} columns, the number represents the number of sample pairs.
    Bold numbers indicate the best scores among different models, while underlined numbers represent the second-best scores.
    }
    \label{tab_main_result}
\end{table*}

\section{Experiment}

\subsection{Experimental Settings}
\label{sec_exp_setup}

\paragraph{Models.}
For open-source multimodal LLMs, we use four variants of InternVL3\_5 (1B, 8B, 14B, and 38B)~\cite{wang2025internvl35advancingopensourcemultimodal}; three variants of Qwen3-VL (4B, 8B, and 30B-A3B)~\cite{yang2025qwen3technicalreport}; two variants of LLaVA-v1.6 (llava-v1.6-mistral-7b and llava-v1.6-vicuna-13b)~\cite{li2024llava}; and Llama-3.2-11B-Vision~\cite{grattafiori2024llama3herdmodels}. We note that the instruction-tuned versions of these models are used.
For the proprietary model, we use OpenAI o4-mini~\cite{openai_o3_operator_2025}.

\paragraph{Prompting Strategies.}
Following the SciVer dataset~\cite{wang-etal-2025-sciver}, we also employ zero-shot Chain-of-Thought~\cite[CoT;][]{wei2022chainofthought} prompting in our experiments.
As shown in Section~\ref{sec_dataset_construction} and Table~\ref{tab_dataset_stats}, our dataset includes context usage information. \textbf{No} indicates that only the claim and a figure or table are needed. \textbf{Yes} indicates that a short context, consisting of the sentences preceding the claim, is required. \textbf{Full paper} indicates that the full text may be needed. Based on this, we design two setups: \textit{no-context} for samples not requiring context and \textit{use-context} for samples requiring either a short context or the full paper.

\paragraph{Evaluation.}
Following previous work on the scientific claim verification task~\cite{lu-etal-2023-scitab,ho-etal-2025-table}, we use the macro-F1 evaluation metric in our experiments.
\update{
However, since the task is a binary classification problem (Supported vs. Refuted), macro-F1 alone may be insufficient. A model may achieve a reasonable macro-F1 score through lucky guesses or by exploiting reasoning shortcuts or dataset biases.
To mitigate this effect, we introduce a new evaluation metric, Pair Accuracy, defined as the number of correctly predicted pairs divided by the total number of pairs. A pair is considered correct only if both samples associated with the same claim, one Supported and one Refuted, are predicted correctly.
This metric is more stringent: a robust model unaffected by biases should be able to correctly predict the labels for both evidence files corresponding to the same claim. Notably, while the random baseline for macro-F1 is 0.5, the random baseline for Pair Accuracy is 0.25, making Pair Accuracy a stricter and more discriminative metric than macro-F1.
}


\begin{table}[htb!]
    \centering
    \resizebox{\columnwidth}{!}{%
    \begin{tabular}{l | rr | rr}
    \toprule
    \multirow{2}{*}{\textbf{Model}} &
    
    \multicolumn{2}{c}{\textbf{Validation}} &
    \multicolumn{2}{c}{\textbf{Test}}  \\ 

    \cmidrule(rl){2-3}
    \cmidrule(rl){4-5}


    ~ & F1 & P-Acc & F1 & P-Acc \\\midrule


    \multicolumn{5}{c}{\textbf{Table Evidence}}  \\ \midrule

        \#Samples & 482 & 236 & 523 & 256 \\ 

        \midrule
        InternVL3\_5-8B & 70.6 & 44.5 & 68.1 & 43.8 \\ 
        
        InternVL3\_5-14B & 72.0 & 49.2 & 70.4 & 45.7 \\ 
        
        InternVL3\_5-38B & 71.7 & 47.5 & 72.0 & 48.4 \\ 

        \midrule
        Qwen3-VL-4B & 74.2 & 49.2 & 73.2 & 46.9 \\ 
        
        Qwen3-VL-8B & 75.4 & 51.7 & 72.0 & 48.0 \\ 
        
        Qwen3-VL-30B-A3B & 80.6 & 62.7 & 74.6 & 52.0 \\ 

        \midrule
        o4-mini & 85.4 & 72.5 & 81.7 & 65.6 \\

        
        

        
        


\midrule
    \multicolumn{5}{c}{\textbf{Figure Evidence}}  \\ \midrule

        \#Samples & 265 & 116 & 394 & 180 \\ 

        \midrule
        InternVL3\_5-8B & 52.2 & 16.4 & 54.3 & 21.7 \\ 
        
        InternVL3\_5-14B & 59.7 & 25.9 & 56.5 & 22.8 \\ 
        
        InternVL3\_5-38B & 57.9 & 25.0 & 62.3 & 32.8 \\ 

        \midrule
        Qwen3-VL-4B & 63.7 & 32.8 & 64.9 & 36.7 \\ 
        
        Qwen3-VL-8B & 65.8 & 37.1 & 67.5 & 38.9 \\ 
        
        Qwen3-VL-30B-A3B & 66.6 & 38.8 & 66.3 & 38.3 \\ 

        \midrule
        o4-mini & 78.1 & 59.5 & 75.4 & 53.3 \\

        
        

        
        


\bottomrule

\end{tabular}
    }
    \caption{
    Detailed average macro-F1 and pair accuracy scores of the models on our dataset, shown separately for the two types of evidence: tables and figures.
    }
    \label{tab_detailed_result}
\end{table}

\begin{table}[htb!]
    \centering
    \resizebox{\columnwidth}{!}{%
    \begin{tabular}{l | rr | rr}
    \toprule
    \multirow{2}{*}{\textbf{Model}} &
    
    \multicolumn{2}{c}{\textbf{Validation}} &
    \multicolumn{2}{c}{\textbf{Test}}  \\ 

    \cmidrule(rl){2-3}
    \cmidrule(rl){4-5}


    ~ & F1 & P-Acc & F1 & P-Acc \\


\midrule
    \multicolumn{5}{c}{\textbf{Graph Flip}}  \\ \midrule

    \#Samples & 40 & 20 & 40 & 20 \\ \midrule

Qwen3-VL-8B & 69.9 & 45.0 & 75.9 & 55.0 \\ 

Qwen3-VL-30B-A3B & 62.3 & 35.0 & 72.3 & 45.0 \\ 

o4-mini & 79.8 & 60.0 & 82.9 & 65.0 \\

    
    

\midrule

\multicolumn{5}{c}{\textbf{Legend Swap}}  \\ \midrule

    \#Samples & 126 & 63 & 80 & 40 \\ \midrule

Qwen3-VL-8B & 65.5 & 38.1 & 68.4 & 42.5 \\ 

Qwen3-VL-30B-A3B & 66.9 & 39.7 & 55.3 & 27.5 \\ 

o4-mini & 81.6 & 63.5 & 79.2 & 57.5 \\

    
    

\midrule
    \multicolumn{5}{c}{\textbf{Graph Swap}}  \\ \midrule
 
    \#Samples & 20 & 10 & 18 & 9 \\ \midrule

Qwen3-VL-8B & 56.3 & 20.0 & 45.2 & 11.1 \\

Qwen3-VL-30B-A3B & 71.4 & 40.0 & 66.2 & 33.3 \\ 

o4-mini & 76.8 & 50.0 & 68.1 & 44.4 \\

    
    

\midrule
    \multicolumn{5}{c}{\textbf{Category Swap}}  \\ \midrule
    
\#Samples & 46 & 23 & 26 & 13 \\ \midrule

Qwen3-VL-8B & 65.0 & 34.8 & 70.6 & 46.2 \\ 

Qwen3-VL-30B-A3B & 66.9 & 39.1 & 82.4 & 61.5 \\ 

o4-mini & 70.3 & 52.2 & 72.1 & 46.2 \\




\midrule
    \multicolumn{5}{c}{\textbf{Others}}  \\ \midrule

\#Samples & 0 & 0 & 196 & 98 \\  \midrule

Qwen3-VL-8B & - & - & 65.8 &	35.7 \\ 

Qwen3-VL-30B-A3B & - & - & 66.4 &	38.8 \\ 

o4-mini & - & - & 74.0 &	51.0 \\

\midrule
    \multicolumn{5}{c}{\textbf{Supported Claim Only}}  \\ \midrule

\#Samples & 33 & ~ & 34 & ~ \\ \midrule

Qwen3-VL-8B & 41.1 & ~ & 41.4 & ~ \\ 

Qwen3-VL-30B-A3B & 42.1 & ~ & 43.3 & ~ \\

o4-mini & 43.1 & ~ & 41.4 & ~ \\

\bottomrule

\end{tabular}
    }
    \caption{
    Detailed average macro-F1 and pair accuracy scores of the models using figure-based evidence are shown separately for different types of evidence modification operations.
    }
    \label{tab_detailed_result_ope}
\end{table}

\subsection{Results}
\label{sec_results}

Table~\ref{tab_main_result} presents the macro-F1 and pair accuracy of the models on our dataset.

\paragraph{Macro-F1 vs. Pair Accuracy.}
\update{
As shown in the table, Pair Accuracy scores are consistently lower than macro-F1 scores across all cases. This indicates that Pair Accuracy is a stricter evaluation metric, which helps reduce inflated performance caused by model guessing.
Specifically, if a model correctly predicts the label for either the Supported or Refuted sample but fails on the other sample within the same claim pair, despite only slight changes in the evidence file, this suggests that the model does not truly understand the evidence. Instead, its predictions may rely on superficial or spurious features rather than genuine reasoning over the evidence.
}

\paragraph{Using No Context vs. Context.}
Based on the prompt strategies in Section~\ref{sec_exp_setup}, we evaluate each validation and test subset using two setups: no-context and use-context. 
\update{
As shown in the table, in most cases, models perform better without additional context than with context, with the no-context setting often achieving higher scores than settings that require context. Overall, these results suggest that solving the task without context is generally easier, although additional context can be beneficial in some cases.
Since both tables and figures already include contextual information in their captions, providing extra context is not always helpful. When the task requires the model to jointly reason over both the context and the evidence, performance tends to decrease, indicating increased difficulty. In contrast, when context serves as supplementary information that does not require complex reasoning, it can improve performance.
}

\paragraph{Open-source vs. Proprietary Models.}
As shown in Table~\ref{tab_main_result}, o4-mini outperforms all open-source MLLMs across all settings, highlighting the performance gap that remains between proprietary and open-source models. 
Among open-source MLLMs, Qwen3-VL-30B-A3B achieves the best results in most cases, while Qwen3-VL-8B leads in the use-context test setting.
As recent models, the Qwen3-VL series demonstrate clear progress in open-source MLLM development.

\subsection{Analyses}
\label{sec_analyses}

To explore model performance in depth, we analyze predictions from several top-performing models. First, we compare samples using tables versus figures as evidence. Second, we examine the effect of evidence modifications on model behavior.

\paragraph{Table Evidence vs. Figure Evidence.}
Table~\ref{tab_detailed_result} shows the detailed average macro-F1 and pair accuracy scores of the models on our dataset, shown separately for the two types of evidence: tables and figures.
As shown in the table, the results of all models on table-based evidence are higher than those on figure-based evidence. This suggests that samples with table evidence are less challenging for the models, whereas samples with figure evidence are more difficult. 
For example, on o4-mini, the validation score for table-evidence samples is 85.4 macro-F1, compared to only 78.1 for figure-evidence samples.

\begin{table}[htb!]
    \centering
    \resizebox{\columnwidth}{!}{%
    \begin{tabular}{l | rr | rr}
    \toprule
    \multirow{2}{*}{\textbf{Model}} &
    
    \multicolumn{2}{c}{\textbf{Validation}} &
    \multicolumn{2}{c}{\textbf{Test}}  \\ 

    \cmidrule(rl){2-3}
    \cmidrule(rl){4-5}


    ~ & F1 & P-Acc & F1 & P-Acc \\\midrule


    \multicolumn{5}{c}{\textbf{Table Evidence}}  \\ \midrule

\#Samples & 472 & 236 & 512 & 256 \\ \midrule

Qwen3-VL-8B & 75.2 & 51.7 & 72.7 & 48.0 \\ 

Qwen3-VL-30B-A3B & 80.6 & 62.7 & 74.9 & 52.0 \\ 

o4-mini & 85.5 & 72.5 & 82.3 & 65.6 \\




\midrule
    \multicolumn{5}{c}{\textbf{Figure Evidence}}  \\ \midrule

\#Samples & 232 & 116 & 360 & 180 \\ \midrule

Qwen3-VL-8B & 65.4 & 37.1 & 67.1 & 38.9 \\ 

Qwen3-VL-30B-A3B & 66.6 & 38.8 & 65.9 & 38.3 \\ 

o4-mini & 78.7 & 59.5 & 75.8 & 53.3 \\




\bottomrule

\end{tabular}
    }
    \caption{
    Detailed average macro-F1 and pair accuracy scores of the models on our dataset, shown separately for the two types of evidence: tables and figures.
    We exclude all samples that contain a note indicating the claim is supported only.
    }
    \label{tab_no_supported_only}
\end{table}

\paragraph{Figure Evidence Modification Operations.}

Based on the previous results, we focus on samples that use figures as evidence. 
Table~\ref{tab_detailed_result_ope} shows the models' average macro-F1 and pair accuracy scores for each type of figure evidence modification.

As discussed in Section~\ref{subsec_data_statistics}, there are 67 ``supported claim only'' samples in the figure-based evidence subset. These samples have very low scores, likely because they are difficult even for annotators to verify and create unsupported claims, making them a hard subset. To investigate whether this explains the models’ poorer performance on figure-based evidence compared to table-based evidence, we recalculated results after removing all ``supported claim only'' samples. The results, shown in Table~\ref{tab_no_supported_only}, confirm that reasoning over figures remains more challenging than over tables.
To illustrate how ``supported claim only'' samples affect table-based evidence results, we show detailed scores for different operations in Table~\ref{tab_detail_ope_tab} in Appendix~\ref{app_analysis}.

Excluding the ``supported claim only'' subset in Table~\ref{tab_detailed_result_ope}, we observe that different operations vary in difficulty for the models.
In general, Others, Graph Swap, and Category Swap are more challenging than Graph Flip. 
%

\subsection{Discussion}
Considering the human scores reported in Section~\ref{subsec_data_validate}, the main results in Section~\ref{sec_results}, and the detailed analyses in Section~\ref{sec_analyses}, we demonstrate that the figure-based subset of our dataset is challenging for all evaluated models, including o4-mini, as a large gap remains between the best-performing model and the human baseline.

For the table-based subset, although the macro-F1 score is close to the human baseline, there is a notable performance drop when evaluated using pair accuracy, indicating that there is still room for improvement.
Moreover, while our table evidence data covers a wide range of table formats, this paper focuses exclusively on the .png format. Consequently, substantial research opportunities remain for exploring more diverse table representations within our dataset.

\section{Conclusion}

In this paper, we introduced SciClaimEval, a dataset for scientific claim verification featuring authentic claims, evidence-based negative examples, and diverse data formats. SciClaimEval bridges a key gap between synthetic benchmarks and real-world scientific reasoning. Our evaluation of multiple MLLMs shows that the figure-based subset remains challenging for all models, including o4-mini, with a substantial gap from human performance. In contrast, the table-based subset is more suitable for evaluating open-source MLLMs, as o4-mini achieves near-human performance. Moreover, our table data support multiple formats, providing a valuable resource for further research on scientific paper processing. 
We hope SciClaimEval will inspire future work on multimodal understanding and the development of more capable and trustworthy scientific reasoning models.

\section*{Limitations}

Our research has three main limitations.

First, as described in Section~\ref{subsec_data_statistics}, we used GPT-5-nano to convert LaTeX and HTML tables into JSON. Human evaluation on 100 random samples revealed 19 cases with major issues and 22 with minor issues, which may affect the quality of the JSON format. We plan to randomly select a subset for human correction in the future.

Second, our dataset contains more table-based than figure-based evidence samples. However, the o4-mini model’s performance on table-based samples is close to the human baseline, making this subset less challenging.

Third, we only use PNG images for table evidence, leaving other formats unutilized in this work.


\section*{Ethical Statement and Broader Impact}

For PeerJ papers, our dataset included 47 papers licensed under CC BY 4.0 and 2 papers licensed under CC BY-NC 4.0. To ensure proper attribution in accordance with these licenses, we added license information, the paper URL, author names, and the paper title to each paper file in JSON format.

For the ML papers, our dataset includes 39 papers licensed under CC BY 4.0, 8 papers licensed under CC BY-NC-SA 4.0, 2 papers licensed under CC BY-SA 4.0, and 2 papers in the public domain. To ensure proper attribution in accordance with these licenses, we added license information, paper URLs, author names, and paper titles to each paper file in JSON format.

For the NLP domain, papers were collected from the ACL Anthology and manually mapped to their arXiv versions when available. Among these, 32 papers are licensed under the arXiv Non-exclusive Distribution License and 4 papers are licensed under CC BY-NC-ND 4.0. Because these licenses do not meet our intended usage requirements, we instead used the ACL versions of these papers, which are licensed under CC BY 4.0. 
For the remaining papers, we include the arXiv license information (CC BY 4.0: 35 papers; CC BY-NC-SA 4.0: 7 papers; CC BY-SA 4.0: 2 papers).

There are a total of 11 annotators involved in the creation of our dataset. All of them are graduate students or AI/NLP researchers. We do not collect or include any personal or sensitive information in the dataset. Annotators are provided with a detailed guideline during the annotation process. In cases where the guidelines are unclear or ambiguous, they are allowed to provide feedback to the authors of the papers to establish a consistent approach for handling such cases.

\section*{Use of LLMs}
We use ChatGPT and GPT-5 to help verify grammar and enhance the quality of our writing. Most of the initial content, however, is authored by us. All suggestions provided by the models are manually reviewed to ensure they accurately convey the intended information. Additionally, we use GitHub Copilot to assist with the coding process.


\section*{Acknowledgements}
This work was supported by JSPS KAKENHI Grant Number 24K03231,
AID-CNRS NaviTerm project (convention 2022 65 0079 CNRS Occitanie Ouest),
and by the Deutsche Forschungsgemeinschaft (DFG, German Research Foundation) -- 554559555.

\section*{Detailed Analysis}
\label{app_analysis}

Table~\ref{tab_detail_ope_tab} shows the detailed average macro-F1 and pair accuracy scores of models using table-based evidence, reported separately for different types of evidence modification operations.

\begin{table}[htb!]
    \centering
    \resizebox{\columnwidth}{!}{%
    \begin{tabular}{l | rr | rr}
    \toprule
    \multirow{2}{*}{\textbf{Model}} &
    
    \multicolumn{2}{c}{\textbf{Validation}} &
    \multicolumn{2}{c}{\textbf{Test}}  \\ 

    \cmidrule(rl){2-3}
    \cmidrule(rl){4-5}

    ~ & F1 & P-Acc & F1 & P-Acc \\\midrule

    \multicolumn{5}{c}{\textbf{Change the Cell Values}}  \\ \midrule

\#Samples & 366 & 183 & 252 & 126 \\ \midrule

Qwen3-VL-8B & 75.6 & 53.0 & 74.7 & 50.8 \\ 

Qwen3-VL-30B-A3B & 80.7 & 62.8 & 75.7 & 54.8 \\ 

o4-mini & 86.5 & 73.2 & 83.4 & 68.3 \\




\midrule
    \multicolumn{5}{c}{\textbf{Swap Rows or Columns }}  \\ \midrule

\#Samples & 106 & 53 & 222 & 111 \\ \midrule

Qwen3-VL-8B & 73.6 & 47.2 & 71.9 & 46.8 \\ 

Qwen3-VL-30B-A3B & 80.3 & 62.3 & 74.1 & 49.5 \\ 

o4-mini & 82.0 & 69.8 & 82.5 & 65.8 \\




\midrule
    \multicolumn{5}{c}{\textbf{Alter the Tables}}  \\ \midrule

    \#Samples & 0 & 0 & 28 & 14 \\ \midrule

Qwen3-VL-8B & - & - & 56.0 & 28.6 \\ 

Qwen3-VL-30B-A3B & - & - & 72.0 & 42.9 \\ 

o4-mini & - & - & 75.0 & 50.0 \\

\midrule
    \multicolumn{5}{c}{\textbf{Others}}  \\ \midrule

\#Samples & 0 & 0 & 10 & 5 \\ \midrule

Qwen3-VL-8B & - & - & 80.0 & 60.0 \\ 

Qwen3-VL-30B-A3B & - & - & 80.0 & 60.0 \\ 

o4-mini & - & - & 69.7 & 40.0 \\

\midrule
    \multicolumn{5}{c}{\textbf{Supported Claim Only}}  \\ \midrule

\#Samples & 10 & ~ & 11 & ~ \\ \midrule
Qwen3-VL-8B & 44.4 & ~ & 26.7  & ~ \\ 

Qwen3-VL-30B-A3B & 44.4 & ~ & 35.3 & ~ \\ 

o4-mini & 44.4 & ~ & 35.3 & ~ \\

\bottomrule

\end{tabular}
    }
    \caption{
Detailed average macro-F1 and pair accuracy scores of the models using table-based evidence are shown separately for different types of evidence modification operations.
    }
    \label{tab_detail_ope_tab}
\end{table}

\nocite{*}
\section{Bibliographical References}\label{sec:reference}

\bibliographystyle{lrec2026-natbib}
\bibliography{custom}

@inproceedings{lu-etal-2023-scitab,
    title = "{SCITAB}: A Challenging Benchmark for Compositional Reasoning and Claim Verification on Scientific Tables",
    author = "Lu, Xinyuan  and
      Pan, Liangming  and
      Liu, Qian  and
      Nakov, Preslav  and
      Kan, Min-Yen",
    editor = "Bouamor, Houda  and
      Pino, Juan  and
      Bali, Kalika",
    booktitle = "Proceedings of the 2023 Conference on Empirical Methods in Natural Language Processing",
    month = dec,
    year = "2023",
    address = "Singapore",
    publisher = "Association for Computational Linguistics",
    url = "https://aclanthology.org/2023.emnlp-main.483/",
    doi = "10.18653/v1/2023.emnlp-main.483",
    pages = "7787--7813",
}

@inproceedings{wang-etal-2021-semeval,
    title = "{S}em{E}val-2021 Task 9: Fact Verification and Evidence Finding for Tabular Data in Scientific Documents ({SEM}-{TAB}-{FACTS})",
    author = "Wang, Nancy X. R.  and
      Mahajan, Diwakar  and
      Danilevsky, Marina  and
      Rosenthal, Sara",
    editor = "Palmer, Alexis  and
      Schneider, Nathan  and
      Schluter, Natalie  and
      Emerson, Guy  and
      Herbelot, Aurelie  and
      Zhu, Xiaodan",
    booktitle = "Proceedings of the 15th International Workshop on Semantic Evaluation (SemEval-2021)",
    month = aug,
    year = "2021",
    address = "Online",
    publisher = "Association for Computational Linguistics",
    url = "https://aclanthology.org/2021.semeval-1.39/",
    doi = "10.18653/v1/2021.semeval-1.39",
    pages = "317--326",
}

@inproceedings{wang-etal-2025-sciver,
    title = "{S}ci{V}er: Evaluating Foundation Models for Multimodal Scientific Claim Verification",
    author = "Wang, Chengye  and
      Shen, Yifei  and
      Kuang, Zexi  and
      Cohan, Arman  and
      Zhao, Yilun",
    editor = "Che, Wanxiang  and
      Nabende, Joyce  and
      Shutova, Ekaterina  and
      Pilehvar, Mohammad Taher",
    booktitle = "Proceedings of the 63rd Annual Meeting of the Association for Computational Linguistics (Volume 1: Long Papers)",
    month = jul,
    year = "2025",
    address = "Vienna, Austria",
    publisher = "Association for Computational Linguistics",
    url = "https://aclanthology.org/2025.acl-long.420/",
    doi = "10.18653/v1/2025.acl-long.420",
    pages = "8562--8579",
    ISBN = "979-8-89176-251-0",
}

@article{zhang2025atomicreasoningscientifictable,
      title={Atomic Reasoning for Scientific Table Claim Verification}, 
      author={Yuji Zhang and Qingyun Wang and Cheng Qian and Jiateng Liu and Chenkai Sun and Denghui Zhang and Tarek Abdelzaher and Chengxiang Zhai and Preslav Nakov and Heng Ji},
      year={2025},
      eprint={2506.06972},
      journal={arXiv:2506.06972},
      primaryClass={cs.CL},
      url={https://arxiv.org/abs/2506.06972}, 
}

@inproceedings{wei2022chainofthought,
 author = {Wei, Jason and Wang, Xuezhi and Schuurmans, Dale and Bosma, Maarten and ichter, brian and Xia, Fei and Chi, Ed and Le, Quoc V and Zhou, Denny},
 booktitle = {Advances in Neural Information Processing Systems},
 editor = {S. Koyejo and S. Mohamed and A. Agarwal and D. Belgrave and K. Cho and A. Oh},
 pages = {24824--24837},
 publisher = {Curran Associates, Inc.},
 title = {Chain-of-Thought Prompting Elicits Reasoning in Large Language Models},
 url = {https://proceedings.neurips.cc/paper_files/paper/2022/file/9d5609613524ecf4f15af0f7b31abca4-Paper-Conference.pdf},
 volume = {35},
 year = {2022}
}

@online{meta2025llama4,
  author       = {Meta-AI},
  title        = {“The Llama 4 Herd: The Beginning of a New Era of Natively Multimodal Intelligence”},
  year         = {2025},
  month        = {apr},
  day          = {5},
  url          = {https://ai.meta.com/blog/llama-4-multimodal-intelligence/},
}

@article{li2024llava,
  	title={LLaVA-OneVision: Easy Visual Task Transfer},
  	author={Li, Bo and Zhang, Yuanhan and Guo, Dong and Zhang, Renrui and Li, Feng and Zhang, Hao and Zhang, Kaichen and Li, Yanwei and Liu, Ziwei and Li, Chunyuan},
  	journal={arXiv:2408.03326},
  	year={2024}
}

@article{yang2025qwen3technicalreport,
      title={Qwen3 Technical Report}, 
      author={An Yang and Anfeng Li and Baosong Yang and Beichen Zhang and Binyuan Hui and Bo Zheng and Bowen Yu and Chang Gao and Chengen Huang and Chenxu Lv and Chujie Zheng and Dayiheng Liu and Fan Zhou and Fei Huang and Feng Hu and Hao Ge and Haoran Wei and Huan Lin and Jialong Tang and Jian Yang and Jianhong Tu and Jianwei Zhang and Jianxin Yang and Jiaxi Yang and Jing Zhou and Jingren Zhou and Junyang Lin and Kai Dang and Keqin Bao and Kexin Yang and Le Yu and Lianghao Deng and Mei Li and Mingfeng Xue and Mingze Li and Pei Zhang and Peng Wang and Qin Zhu and Rui Men and Ruize Gao and Shixuan Liu and Shuang Luo and Tianhao Li and Tianyi Tang and Wenbiao Yin and Xingzhang Ren and Xinyu Wang and Xinyu Zhang and Xuancheng Ren and Yang Fan and Yang Su and Yichang Zhang and Yinger Zhang and Yu Wan and Yuqiong Liu and Zekun Wang and Zeyu Cui and Zhenru Zhang and Zhipeng Zhou and Zihan Qiu},
      year={2025},
      eprint={2505.09388},
      journal={arXiv:2505.09388},
      primaryClass={cs.CL},
      url={https://arxiv.org/abs/2505.09388}, 
}

@article{wang2025internvl35advancingopensourcemultimodal,
      title={InternVL3.5: Advancing Open-Source Multimodal Models in Versatility, Reasoning, and Efficiency}, 
      author={Weiyun Wang and Zhangwei Gao and Lixin Gu and Hengjun Pu and Long Cui and Xingguang Wei and Zhaoyang Liu and Linglin Jing and Shenglong Ye and Jie Shao and Zhaokai Wang and Zhe Chen and Hongjie Zhang and Ganlin Yang and Haomin Wang and Qi Wei and Jinhui Yin and Wenhao Li and Erfei Cui and Guanzhou Chen and Zichen Ding and Changyao Tian and Zhenyu Wu and Jingjing Xie and Zehao Li and Bowen Yang and Yuchen Duan and Xuehui Wang and Zhi Hou and Haoran Hao and Tianyi Zhang and Songze Li and Xiangyu Zhao and Haodong Duan and Nianchen Deng and Bin Fu and Yinan He and Yi Wang and Conghui He and Botian Shi and Junjun He and Yingtong Xiong and Han Lv and Lijun Wu and Wenqi Shao and Kaipeng Zhang and Huipeng Deng and Biqing Qi and Jiaye Ge and Qipeng Guo and Wenwei Zhang and Songyang Zhang and Maosong Cao and Junyao Lin and Kexian Tang and Jianfei Gao and Haian Huang and Yuzhe Gu and Chengqi Lyu and Huanze Tang and Rui Wang and Haijun Lv and Wanli Ouyang and Limin Wang and Min Dou and Xizhou Zhu and Tong Lu and Dahua Lin and Jifeng Dai and Weijie Su and Bowen Zhou and Kai Chen and Yu Qiao and Wenhai Wang and Gen Luo},
      year={2025},
      eprint={2508.18265},
      journal={arXiv:2508.18265},
      primaryClass={cs.CV},
      url={https://arxiv.org/abs/2508.18265}, 
}

@article{yang2025doestablesourcematter,
      title={Does Table Source Matter? Benchmarking and Improving Multimodal Scientific Table Understanding and Reasoning}, 
      author={Bohao Yang and Yingji Zhang and Dong Liu and André Freitas and Chenghua Lin},
      year={2025},
      eprint={2501.13042},
      journal={arXiv:2501.13042},
      primaryClass={cs.CL},
      url={https://arxiv.org/abs/2501.13042}, 
}

@inproceedings{thorne-etal-2018-fever,
    title = "{FEVER}: a Large-scale Dataset for Fact Extraction and {VER}ification",
    author = "Thorne, James  and
      Vlachos, Andreas  and
      Christodoulopoulos, Christos  and
      Mittal, Arpit",
    editor = "Walker, Marilyn  and
      Ji, Heng  and
      Stent, Amanda",
    booktitle = "Proceedings of the 2018 Conference of the North {A}merican Chapter of the Association for Computational Linguistics: Human Language Technologies, Volume 1 (Long Papers)",
    month = jun,
    year = "2018",
    address = "New Orleans, Louisiana",
    url = "https://aclanthology.org/N18-1074/",
    doi = "10.18653/v1/N18-1074",
    pages = "809--819",
}

@inproceedings{wang-2017-liar,
    title = "``Liar, Liar Pants on Fire'': A New Benchmark Dataset for Fake News Detection",
    author = "Wang, William Yang",
    editor = "Barzilay, Regina  and
      Kan, Min-Yen",
    booktitle = "Proceedings of the 55th Annual Meeting of the Association for Computational Linguistics (Volume 2: Short Papers)",
    month = jul,
    year = "2017",
    address = "Vancouver, Canada",
    url = "https://aclanthology.org/P17-2067/",
    doi = "10.18653/v1/P17-2067",
    pages = "422--426",
}

@inproceedings{jiang-etal-2020-hover,
    title = "{H}o{V}er: A Dataset for Many-Hop Fact Extraction And Claim Verification",
    author = "Jiang, Yichen  and
      Bordia, Shikha  and
      Zhong, Zheng  and
      Dognin, Charles  and
      Singh, Maneesh  and
      Bansal, Mohit",
    editor = "Cohn, Trevor  and
      He, Yulan  and
      Liu, Yang",
    booktitle = "Findings of the Association for Computational Linguistics: EMNLP 2020",
    month = nov,
    year = "2020",
    address = "Online",
    url = "https://aclanthology.org/2020.findings-emnlp.309/",
    doi = "10.18653/v1/2020.findings-emnlp.309",
    pages = "3441--3460",
}

@inproceedings{wadden-etal-2020-fact,
    title = "Fact or Fiction: Verifying Scientific Claims",
    author = "Wadden, David  and
      Lin, Shanchuan  and
      Lo, Kyle  and
      Wang, Lucy Lu  and
      van Zuylen, Madeleine  and
      Cohan, Arman  and
      Hajishirzi, Hannaneh",
    editor = "Webber, Bonnie  and
      Cohn, Trevor  and
      He, Yulan  and
      Liu, Yang",
    booktitle = "Proceedings of the 2020 Conference on Empirical Methods in Natural Language Processing (EMNLP)",
    month = nov,
    year = "2020",
    address = "Online",
    publisher = "Association for Computational Linguistics",
    url = "https://aclanthology.org/2020.emnlp-main.609/",
    doi = "10.18653/v1/2020.emnlp-main.609",
    pages = "7534--7550",
}

@inproceedings{wadden-etal-2022-scifact,
    title = "{S}ci{F}act-Open: Towards open-domain scientific claim verification",
    author = "Wadden, David  and
      Lo, Kyle  and
      Kuehl, Bailey  and
      Cohan, Arman  and
      Beltagy, Iz  and
      Wang, Lucy Lu  and
      Hajishirzi, Hannaneh",
    editor = "Goldberg, Yoav  and
      Kozareva, Zornitsa  and
      Zhang, Yue",
    booktitle = "Findings of the Association for Computational Linguistics: EMNLP 2022",
    month = dec,
    year = "2022",
    address = "Abu Dhabi, United Arab Emirates",
    publisher = "Association for Computational Linguistics",
    url = "https://aclanthology.org/2022.findings-emnlp.347/",
    doi = "10.18653/v1/2022.findings-emnlp.347",
    pages = "4719--4734",
}

@inproceedings{kotonya-toni-2020-explainable-automated,
    title = "Explainable Automated Fact-Checking for Public Health Claims",
    author = "Kotonya, Neema  and
      Toni, Francesca",
    editor = "Webber, Bonnie  and
      Cohn, Trevor  and
      He, Yulan  and
      Liu, Yang",
    booktitle = "Proceedings of the 2020 Conference on Empirical Methods in Natural Language Processing (EMNLP)",
    month = nov,
    year = "2020",
    address = "Online",
    publisher = "Association for Computational Linguistics",
    url = "https://aclanthology.org/2020.emnlp-main.623/",
    doi = "10.18653/v1/2020.emnlp-main.623",
    pages = "7740--7754",
}

@article{guo-etal-2022-survey,
    title = "A Survey on Automated Fact-Checking",
    author = "Guo, Zhijiang  and
      Schlichtkrull, Michael  and
      Vlachos, Andreas",
    editor = "Roark, Brian  and
      Nenkova, Ani",
    journal = "Transactions of the Association for Computational Linguistics",
    volume = "10",
    year = "2022",
    address = "Cambridge, MA",
    publisher = "MIT Press",
    url = "https://aclanthology.org/2022.tacl-1.11/",
    doi = "10.1162/tacl_a_00454",
    pages = "178--206",
}

@inproceedings{aly2021feverous,
    title={{FEVEROUS}: Fact Extraction and {VER}ification Over Unstructured and Structured information},
    author={Rami Aly and Zhijiang Guo and Michael Sejr Schlichtkrull and James Thorne and Andreas Vlachos and Christos Christodoulopoulos and Oana Cocarascu and Arpit Mittal},
    booktitle={Thirty-fifth Conference on Neural Information Processing Systems Datasets and Benchmarks Track (Round 1)},
    year={2021},
    url={https://openreview.net/forum?id=h-flVCIlstW}
}

@inproceedings{augenstein-etal-2019-multifc,
    title = "{M}ulti{FC}: A Real-World Multi-Domain Dataset for Evidence-Based Fact Checking of Claims",
    author = "Augenstein, Isabelle  and
      Lioma, Christina  and
      Wang, Dongsheng  and
      Chaves Lima, Lucas  and
      Hansen, Casper  and
      Hansen, Christian  and
      Simonsen, Jakob Grue",
    editor = "Inui, Kentaro  and
      Jiang, Jing  and
      Ng, Vincent  and
      Wan, Xiaojun",
    booktitle = "Proceedings of the 2019 Conference on Empirical Methods in Natural Language Processing and the 9th International Joint Conference on Natural Language Processing (EMNLP-IJCNLP)",
    month = nov,
    year = "2019",
    address = "Hong Kong, China",
    publisher = "Association for Computational Linguistics",
    url = "https://aclanthology.org/D19-1475/",
    doi = "10.18653/v1/D19-1475",
    pages = "4685--4697",
}

@misc{wkhtmltopdf,
  author       = {Ashish Kulkarni },
  title        = {wkhtmltopdf},
  year         = {2023},
  url          = {https://github.com/wkhtmltopdf/wkhtmltopdf}
}

@misc{belval_pdf2image,
  author       = {Edouard Belval },
  title        = {pdf2image},
  year         = {2023},
  url          = {https://github.com/Belval/pdf2image}
}

@inproceedings{ho-etal-2025-table,
    title = "Table-Text Alignment: Explaining Claim Verification Against Tables in Scientific Papers",
    author = "Ho, Xanh  and
      Kumar, Sunisth  and
      Wu, Yun-Ang  and
      Boudin, Florian  and
      Takasu, Atsuhiro  and
      Aizawa, Akiko",
    editor = "Christodoulopoulos, Christos  and
      Chakraborty, Tanmoy  and
      Rose, Carolyn  and
      Peng, Violet",
    booktitle = "Findings of the Association for Computational Linguistics: EMNLP 2025",
    month = nov,
    year = "2025",
    address = "Suzhou, China",
    publisher = "Association for Computational Linguistics",
    url = "https://aclanthology.org/2025.findings-emnlp.135/",
    doi = "10.18653/v1/2025.findings-emnlp.135",
    pages = "2509--2517",
    ISBN = "979-8-89176-335-7",
}

@inproceedings{lal-etal-2025-musciclaims,
    title = "{M}u{S}ci{C}laims: Multimodal Scientific Claim Verification",
    author = "Lal, Yash Kumar  and
      Bandham, Manikanta  and
      Hasan, Mohammad Saqib  and
      Kashi, Apoorva  and
      Koupaee, Mahnaz  and
      Balasubramanian, Niranjan",
    editor = "Inui, Kentaro  and
      Sakti, Sakriani  and
      Wang, Haofen  and
      Wong, Derek F.  and
      Bhattacharyya, Pushpak  and
      Banerjee, Biplab  and
      Ekbal, Asif  and
      Chakraborty, Tanmoy  and
      Singh, Dhirendra Pratap",
    booktitle = "Proceedings of the 14th International Joint Conference on Natural Language Processing and the 4th Conference of the Asia-Pacific Chapter of the Association for Computational Linguistics",
    month = dec,
    year = "2025",
    address = "Mumbai, India",
    publisher = "The Asian Federation of Natural Language Processing and The Association for Computational Linguistics",
    url = "https://aclanthology.org/2025.ijcnlp-long.175/",
    pages = "3285--3307",
    ISBN = "979-8-89176-298-5",
}

@inproceedings{Chen2020TabFact,
title={TabFact: A Large-scale Dataset for Table-based Fact Verification},
author={Wenhu Chen and Hongmin Wang and Jianshu Chen and Yunkai Zhang and Hong Wang and Shiyang Li and Xiyou Zhou and William Yang Wang},
booktitle={International Conference on Learning Representations},
year={2020},
url={https://openreview.net/forum?id=rkeJRhNYDH}
}

@misc{openai_o3_operator_2025,
  author       = {OpenAI},
  title        = {Addendum to OpenAI o3 and o4-mini System Card: OpenAI o3 Operator},
  year         = {2025},
  url          = {https://openai.com/index/o3-o4-mini-system-card-addendum-operator-o3/},
}

@article{grattafiori2024llama3herdmodels,
      title={The Llama 3 Herd of Models}, 
      author={Aaron Grattafiori and Abhimanyu Dubey and Abhinav Jauhri and Abhinav Pandey and Abhishek Kadian and Ahmad Al-Dahle and Aiesha Letman and Akhil Mathur and Alan Schelten and Alex Vaughan and et al.},
      year={2024},
      eprint={2407.21783},
      archivePrefix={arXiv},
      primaryClass={cs.AI},
      journal={arXiv:2407.21783},
      url={https://arxiv.org/abs/2407.21783}, 
}

@inproceedings{kim-etal-2023-factkg,
    title = "{F}act{KG}: Fact Verification via Reasoning on Knowledge Graphs",
    author = "Kim, Jiho  and
      Park, Sungjin  and
      Kwon, Yeonsu  and
      Jo, Yohan  and
      Thorne, James  and
      Choi, Edward",
    editor = "Rogers, Anna  and
      Boyd-Graber, Jordan  and
      Okazaki, Naoaki",
    booktitle = "Proceedings of the 61st Annual Meeting of the Association for Computational Linguistics (Volume 1: Long Papers)",
    month = jul,
    year = "2023",
    address = "Toronto, Canada",
    publisher = "Association for Computational Linguistics",
    url = "https://aclanthology.org/2023.acl-long.895/",
    doi = "10.18653/v1/2023.acl-long.895",
    pages = "16190--16206",
}

@inproceedings{akhtar-etal-2024-chartcheck,
    title = "{C}hart{C}heck: Explainable Fact-Checking over Real-World Chart Images",
    author = "Akhtar, Mubashara  and
      Subedi, Nikesh  and
      Gupta, Vivek  and
      Tahmasebi, Sahar  and
      Cocarascu, Oana  and
      Simperl, Elena",
    editor = "Ku, Lun-Wei  and
      Martins, Andre  and
      Srikumar, Vivek",
    booktitle = "Findings of the Association for Computational Linguistics: ACL 2024",
    month = aug,
    year = "2024",
    address = "Bangkok, Thailand",
    publisher = "Association for Computational Linguistics",
    url = "https://aclanthology.org/2024.findings-acl.828/",
    doi = "10.18653/v1/2024.findings-acl.828",
    pages = "13921--13937",
}

@ARTICLE{wang2004ssim,
  author={Zhou Wang and Bovik, A.C. and Sheikh, H.R. and Simoncelli, E.P.},
  journal={IEEE Transactions on Image Processing}, 
  title={Image quality assessment: from error visibility to structural similarity}, 
  year={2004},
  volume={13},
  number={4},
  pages={600-612},
  doi={10.1109/TIP.2003.819861}}

@online{SML:ar5iv:04:2024,
  author = {Deyan Ginev},
  title = {ar5iv:04.2024 dataset, an HTML5 conversion of arXiv.org},
  url = {https://sigmathling.kwarc.info/resources/ar5iv-dataset-2024/},
  note = {SIGMathLing -- Special Interest Group on Math Linguistics},
  year = {2024} }


\end{document}